\documentclass[fleqn,10pt]{wlscirep}
\usepackage[utf8]{inputenc}
\usepackage[T1]{fontenc}
\usepackage{booktabs}
\usepackage{pgfplotstable}
\pgfplotsset{compat=1.14}
\usepackage{lineno}
\usepackage{graphicx}

\newcolumntype{x}[1]{>{\centering\arraybackslash}p{#1pt}}

\newlength\savewidth

\makeatletter\renewcommand\paragraph{\@startsection{paragraph}{4}{\z@}
  {.5em \@plus1ex \@minus.2ex}{-.5em}{\normalfont\normalsize\bfseries}}\makeatother
  
\usepackage{xcolor}
\newcommand{\green}[1]{\textcolor{black}{#1}}   
\newcommand{\blue}[1]{\textcolor{black}{#1}}  
\title{A Realistic Fish-Habitat Dataset to Evaluate Algorithms for Underwater Visual Analysis}

\author[1*]{Alzayat Saleh}
\author[2,3*]{Issam H. Laradji}
\author[1]{Dmitry A. Konovalov}
\author[1]{Michael Bradley}
\author[3]{David Vazquez}
\author[1]{Marcus Sheaves}

\affil[1]{James Cook University, \textsuperscript{2} University of British Columbia, \textsuperscript{3} Element AI}
\affil[*]{Equal contribution, alzayat.saleh@my.jcu.edu.au, issam.laradji@gmail.com}


\begin{abstract}
\blue{Visual analysis of complex fish habitats is an important step towards sustainable fisheries for human consumption and environmental protection. Deep Learning methods have shown great promise for scene analysis when trained on large-scale datasets. However, current datasets for fish analysis tend to focus on the classification task within constrained, plain environments which do not capture the complexity of underwater fish habitats. To address this limitation, we present DeepFish as a benchmark suite  with a large-scale dataset to train and test methods for several computer vision tasks. The dataset consists of approximately 40 thousand images  collected underwater from 20  \green{habitats in the} marine-environments of tropical Australia. The dataset originally contained only classification labels. Thus, we collected point-level and segmentation labels to have a more  comprehensive fish analysis benchmark. These labels enable models to learn to automatically monitor fish count, identify their locations, and estimate their sizes. Our experiments provide an in-depth analysis of the dataset characteristics, and the performance evaluation of several state-of-the-art approaches based on our benchmark. Although models pre-trained on ImageNet have successfully performed on this benchmark, there is still room for improvement. Therefore, this benchmark serves as a testbed to motivate further development in this challenging domain of underwater computer vision. Code is available at: \url{https://github.com/alzayats/DeepFish}}
\end{abstract}

\begin{document}
\flushbottom
\maketitle
\thispagestyle{empty}

\section{Introduction}
\blue{Monitoring fish \green{in their natural } habitat is an important step towards sustainable fisheries. In the New South Wales state of Australia, for example, fisheries is valued at more than 100 million Australian dollars in 2012-2013 \cite{Hussain2017AnOO}.  \green{Effective monitoring can } provide information about which areas require protection and restoration to maintain healthy fish populations for both human consumption and environmental protection. Having a system that can automatically perform  comprehensive monitoring can significantly reduce  labour costs and increase  efficiency. The system can lead to a large positive sustainability impact and improve our ability to maintain a healthy ecosystem.}

\blue{Deep learning methods have consistently achieved state-of-the-art results \green{in image  analysis}. Many methods based on deep neural networks achieved top performance for a variety of applications, including, ecological monitoring with camera trap data. One reason behind this success is that these methods can leverage  large-scale, publicly available datasets such as ImageNet~\cite{imagenet_cvpr09} and COCO~\cite{lin2014microsoft} for training before being fine-tuned for a new application.} 

\begin{figure}[t]
\centering
\includegraphics[width=0.99\textwidth]{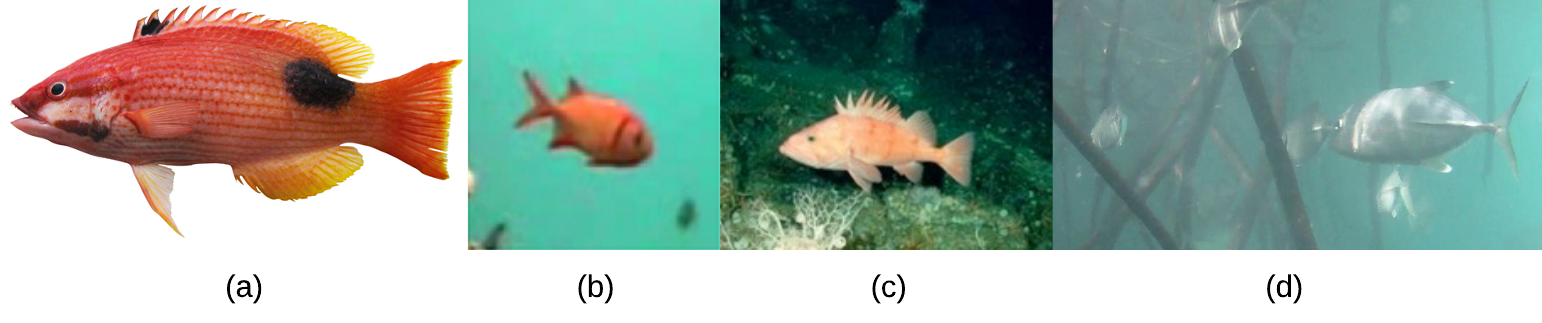}
\caption{\textbf{A comparison of fish datasets.} 
(a) QUT~\cite{anantharajah2014local}, (b) Fish4Knowledge~\cite{f4kFinalReport}, (c) Rockfish~\cite{Rockfish2013}, and (d) our proposed dataset DeepFish. (a-c) datasets are acquired from constrained environments, whereas \textit{DeepFish} has more realistic and challenging environments.
(Figures a-c were obtained from the open-source datasets \cite{anantharajah2014local,f4kFinalReport,Rockfish2013} )}
\label{qutf4k}
\end{figure}

\blue{A particularly challenging application involves automatic analysis of underwater fish habitats which demands a comprehensive, accurate computer vision system. Thus, considerable research efforts have been put towards developing systems for the task of understanding complex marine environments and distinguishing between a diverse set of fish species, which are based on publicly available fish datasets~\cite{LifeCLEF2014, BOOM2014, f4kFinalReport, anantharajah2014local, Rockfish2013}. However, these
fish datasets are small and do not fully capture the variability and complexity of real-world  underwater habitats which often have adversarial water conditions, high similarity of the appearance between fish and some elements in the background such as rocks, and occlusions between fish. For example, the QUT fish dataset~\cite{anantharajah2014local} contains only 3960 labelled images of 468 species. Many of these  fish images are taken in controlled environments where the background is plain white and the illumination is carefully adjusted (Figure~\ref{qutf4k} (a)). Similarly, underwater images collected for the  Fish4Knowledge~\cite{f4kFinalReport} and Rockfish~\cite{Rockfish2013}  datasets are cropped to have a single fish shown at the center (Figure~\ref{qutf4k} (b,c)), which requires costly human labor \green{to produce } and does not help models learn to \green{recognise} fish  in the wild.  Thus, the limitations of these datasets can inhibit further progress in building systems for comprehensive visual understanding of \green{underwater environments}. }

\blue{To this end, we propose {DeepFish} as a benchmark that includes a dataset based on \green{in-situ field recordings of fish habitats} \cite{Bradley2019}  and  we tailor it towards analyzing fish  in underwater marine environments. The dataset consists of approximately 40 thousand  high-resolution ($1920\times 1080$) images collected underwater from 20 \green{different marine habitats in} tropical Australia (see Figure~\ref{qutf4k}~(d) for an example image). These  represent  the  breadth of \green{different } coastal and nearshore benthic habitats commonly available to fish species \green{in tropical Australia} ~\cite{sheaves2009consequences}  (Figure~\ref{fig:areas}). }

\blue{Further, we go beyond the original classification labels  by also acquiring point-level and semantic  segmentation labels for additional computer vision tasks. These labels allow models to learn to analyze fish habitats from several perspectives, including, understanding fish dynamics, monitoring their count, and estimating their sizes and shapes. We evaluate state-of-the-art  methods on these labels to analyze the dataset characteristics and establish initial results for this benchmark.}

\blue{Overall, we can summarize our contributions as follows; (1) we present a benchmark that includes a dataset that captures the complexity and diversity of underwater fish habitats compared to previous fish datasets, (2) we incorporate additional labels to allow for a more comprehensive analysis of fish \green{in underwater environments, } (3) we show  the importance of having pretrained models for achieving good performance on the benchmark, and (4) we provide results that can serve as reference for evaluating new methods. The dataset and the code \green{have been} made public to help spark progress in developing systems for analysing fish habitats.}

\section{Dataset}

Our goal is to design a benchmark that can enable significant progress in fish habitat understanding. Thus we carefully look into the quality of data acquisition,  preparation, and annotation protocol.

Accordingly, we start with the dataset \green{based on the work of Bradley and colleagues} ~\cite{Bradley2019} as it consists of a large number of images (around 40 thousand) that capture high variability of underwater fish habitats. The dataset's diversity and size makes it suitable for training and evaluating deep learning methods. However, the dataset's original purpose was not to evaluate machine learning methods. It was to examine the interactive structuring effects of local habitat characteristics and environmental context on assemblage composition of juvenile fish. 

Yet the characteristics of the dataset makes it suitable as a machine learning benchmark. We tailor it to make the dataset a more comprehensive testbed to spark new, specialized algorithms in this problem setup and name the dataset as {\it DeepFish}. In the following sections we discuss how the data was collected, the additional annotations acquired for the dataset, how it was split between training, validation and testing, and how the dataset compares with current fish datasets.

\subsection{Data collection}
\begin{figure}[t]
\centering
\includegraphics[width=0.49\textwidth]{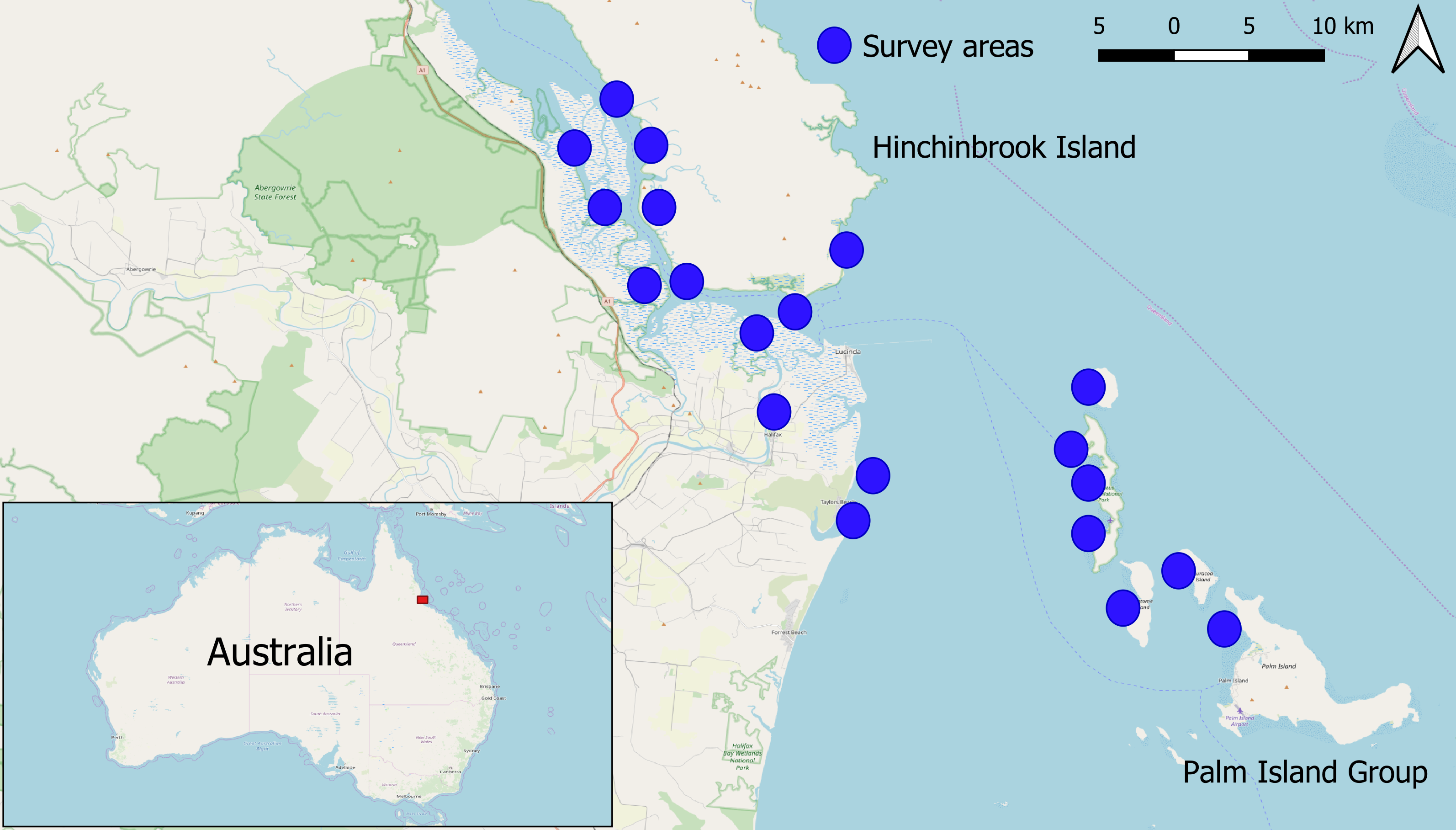}
\caption{\textbf{Locations where the DeepFish images were acquired.} Most of DeepFish has been acquired from the  Hinchinbrook/Palm Islands region in North Eastern Australia, the rest from Western Australia (\textit{not shown in the map}). (The map was created using QGIS  version 3.8, which is available at https://qgis.org)}
\label{fig:areas}
\end{figure}

\begin{figure}[t]
\centering
\includegraphics[width=0.8\textwidth]{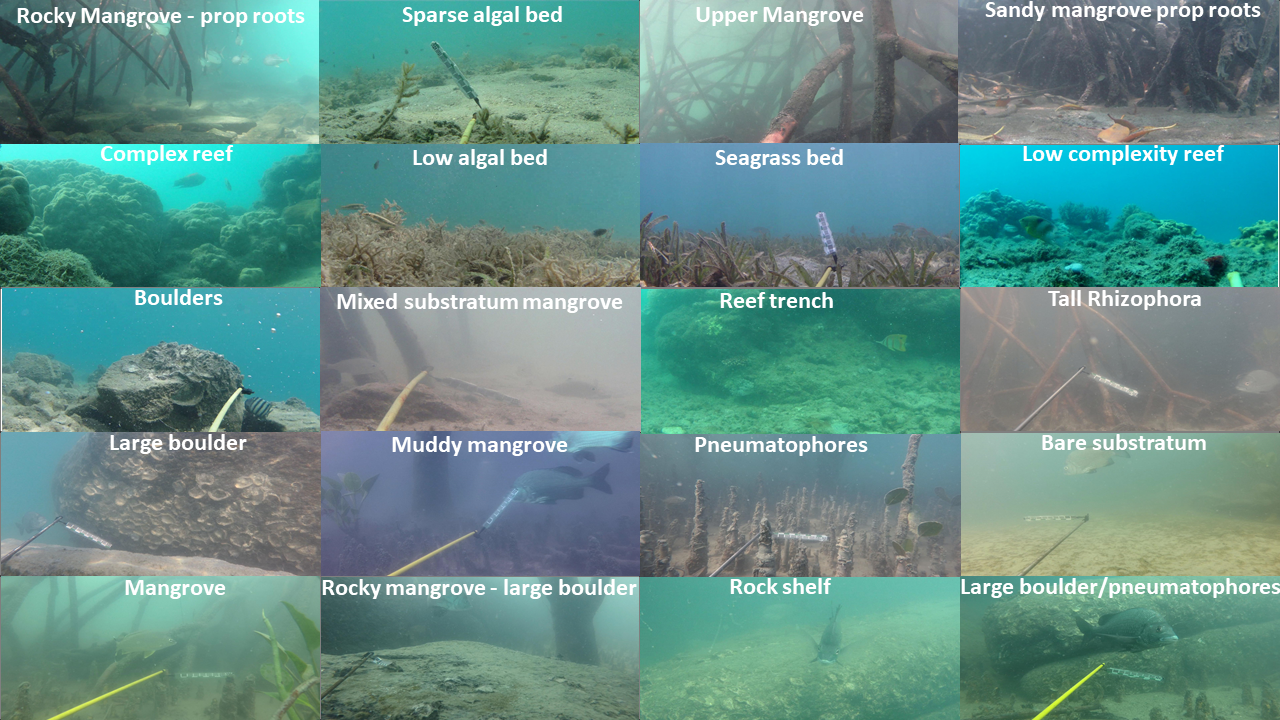}
\caption{DeepFish image samples across 20 different habitats.}
\label{fig:habitat_images}
\end{figure}

Videos for DeepFish were collected for 20 habitats from remote coastal marine environments of tropical Australia (Figure~\ref{fig:areas}). These videos were acquired using cameras mounted on metal frames, deployed over the side of a vessel to acquire video footage underwater. The cameras were lowered to the seabed and left to record the natural fish community, while the vessel maintained a distance of 100m. The depth and the map coordinates of the  cameras were collected using an acoustic depth sounder and a GPS, respectively. Video recording were carried out \green{during} daylight hours, and in relatively low turbidity periods. The video clips were captured in full HD resolution (\textit{1920 x 1080 pixels}) from a digital camera. In total, the number of video frames taken is 39,766 and their distribution across the habitats are shown in Table~\ref{tab:datasets}. Examples of these video frames are shown in Figure~\ref{fig:habitat_images} which illustrate the diversity between the habitats.

This method of acquiring images is a low-disturbance technique that allows us to accurately assess fish-habitats \green{associations} in challenging, even inaccessible  environments~\cite{Bradley2019}. In contrast, many existing monitoring techniques used to understand fish habitats suffer from  the problem of fish flight response, especially for habitats with limited visibility~\cite{sheaves2016use}. For example, a common surveying technique requires divers to conduct visual census~\cite{barnes2012use}, which can cause disturbance to the fish, leading to inaccurate assessment of the fish \green{community}. Furthermore, divers cannot access areas with predators such as crocodiles with this technique. Other surveying techniques involve netting~\cite{sheaves2012importance} and trawling~\cite{rozas1997estimating} for catching and counting fish. However, these methods are invasive and interfere with the behaviour of the fish which can lead to inaccurate estimates. Further, they are limited to estimating fish count only. On the other hand, the data collection procedure for \textit{DeepFish} is one of the most efficient methods for capturing a realistic, unaltered view of fish habitats \green{associations}~\cite{Bradley2019}.

\subsection{Additional Annotations}
The original labels of the dataset are only suitable for the task of classification. These labels were acquired for each video frame, and they indicate whether an image has fish or not (regardless of the count of fish). These labels can be useful to train models for analyzing a fish utilization estimate between different habitats~\cite{Konovalov2019UnderwaterFD}. For example, classifying images between those that contain and do not contain fish allows biologists and ecologists 
to focus their efforts by analyzing only those images with the fish. However, they do not allow for a more detailed analysis of the habitats.

To address this limitation, we acquired point-level and semantic  segmentation labels to enable models to learn to perform the computer vision tasks such as object counting, localization and segmentation. Point-level annotations are provided as a single click on each fish as shown in Figure~\ref{fig:qualitative} (b), and segmentation labels as boundaries over the fish instances Figure~\ref{fig:qualitative}(d) We describe them in detail in the following sections.

\paragraph{Point-level annotations.} The goal of these annotations is to enable models to learn to perform fish counting. A useful application of this task is to automatically monitor fish population in order to avoid the risk of overfishing. These annotations also enable the task of localizing fish within each image which can be used for fish tracking and fish dynamics analysis. 

We annotated 3200 images with point-level annotations which we acquired across different habitats as shown in Table~\ref{tab:statistics}. These annotations represent the $(x, y)$ coordinates of each fish within the images and they are placed around the centroid of the corresponding fish (Figure~\ref{fig:qualitative}(b)). These annotations were acquired using Labelme~\cite{labelme2016}, which is an open-source image annotation tool. It took approximately 1 second per fish with a labeler who is familiar with fish habitats. Since there is an average of 7 fish in each image, the annotation time is estimated at 7 seconds per image. Thus, this labeling scheme makes it easy to acquire additional annotations for new images and fish classes.

\paragraph{Per-pixel annotations.} The goal of these annotations is to train and evaluate models to segment  fish across images. As a result, the segmentation output can be used to estimate fish sizes, shapes, and their weight as shown in~\cite{konovalov2018estimating,konovalov2019automatic}. These are important statistics that can be useful in applications like commercial trawling~\cite{garcia2019automatic}.

We collected per-pixel labels for 620 images. We labeled the fish using layered polygons in order to distinguish between pixels that belong to fish and those to the background (Figure~\ref{fig:qualitative}(c)). The pixel labels represent the size and shape of the fishes in the image. We used Lear~\cite{lear} to extract these segmentation masks, an open-source image annotation tool commonly used for obtaining segmentation labels. Acquiring per-pixel labels is vastly more time-consuming than point-level annotations. It took  around 2 minutes to label a single fish, to ensure quality masks we multiplied the manually generated masks with the original images to visually check the quality of the segmentation. In total, it took around 25 hours to acquire segmentation labels for 310 valid images out of 620 images which is around 5 minutes per image. We acquired labels for a variety of habitats as shown in Table~\ref{tab:statistics}. We see that no point-level nor per-pixel labels were collected for "Sparse algal bed". The reason is that the videos taken for the habitat shows hundreds of tiny fish in each frame where many of them are occluded and are indistinguishable from debris and tiny rocks. As a consequence, it is difficult to annotate a single image for localization and segmentation.

\subsection{Dataset splits}
We define a sub-dataset for each computer vision task: \textit{FishClf} for classification, \textit{FishLoc} for counting and localization, and \textit{FishSeg} for the segmentation task. For each sub-dataset, we split the annotated images into training, validation, and test sets. Instead of splitting the data completely at random, we consider 
each split to represent the variability of different fish habitats and to have similar fish population size. Concretely, we first divide each habitat into images with no fish (background) and images with at least one fish (foreground). We randomly select 50\% training,  20\% for validation and 30\% for testing for each habitat while ensuring that the number of background and foreground images are equal between them. Finally,  we aggregate the selected training images from each habitat into one training split for the dataset. We do the same for the validation and testing splits.

As a result, we get a unique split consisting of 19883, 7953, 11930  (training, validation and test) for \textit{FishClf}, 1600, 640, 960 for \textit{FishLoc}, and 310, 124, 186 for \textit{FishSeg}. While all the annotations, including for the test images, are made available, the expected evaluation setup is to select the best model on the validation set and perform a single evaluation on the test set. The reported results on the test set are then presented in a leaderboard to compare between the algorithms.

\begin{table}[t]
\centering
\caption{{\bf \textit{DeepFish} Dataset Statistics.} Number of images annotated for each sub-dataset: FishClf for classification, FishLoc for counting/localization, and FishSeg for semantic segmentation}
\label{tab:statistics}
\resizebox{0.5\textwidth}{!}{%
\begin{tabular}{l|c|c|c}
\textbf{Habitats} & \textbf{FishClf} & \textbf{FishLoc} & \textbf{FishSeg} \\\hline\hline
Low complexity reef                         & 4977 & 357 & 77 \\ 
Sandy mangrove prop roots                   & 4162 & 322 & 42 \\ 
Complex reef                                & 4018 & 190 & 16 \\ 
Seagrass bed                                & 3255 & 328 & 16 \\ 
Low algal bed                               & 2795 & 282 & 17 \\ 
Reef trench                                 & 2653 & 187 & 48 \\ 
Boulders                                    & 2314 & 227 & 16 \\ 
Mixed substratum mangrove                   & 2139 & 177 & 28 \\ 
Rocky Mangrove - prop roots                 & 2119 & 211 & 27 \\ 
Upper Mangrove                              & 2101 & 129 & 21 \\ 
Rock shelf                                  & 1848 & 186 & 19 \\ 
Mangrove                                    & 1542 & 157 & 33 \\ 
Sparse algal bed                            & 1467 & 0   & 0  \\ 
Muddy mangrove                              & 1117 & 113 & 79 \\ 
Large boulder and pneumatophores            & 900  & 91  & 37 \\ 
Rocky mangrove - large boulder              & 560  & 57  & 28 \\
Bare substratum                             & 526  & 55  & 32 \\
Upper mangrove                              & 475  & 49  & 28 \\
Large boulder                               & 434  & 45  & 27 \\
Muddy mangrove                              & 364  & 37  & 29 \\\hline
\textbf{Total}                          & \textbf{39766} &  \textbf{3200} & \textbf{620}  \\
\end{tabular}}
\end{table}

\subsection{Comparison to other datasets.}

\blue{We compare \textit{DeepFish} to other datasets in terms of (i) dataset size (ii) visual complexity and (iii) vision tasks.  Many  datasets exist for fish analysis~\cite{french2015convolutional, LifeCLEF2014, BOOM2014, f4kFinalReport}. But we chose those that are most similar to ours, namely,  QUT~\cite{anantharajah2014local}, Rockfish~\cite{Rockfish2013}, and Fish4Knowledge~\cite{f4kFinalReport}.}

\blue{Table~\ref{tab:datasets} shows that \textit{DeepFish} is the largest dataset with images of highest resolution. \green{Unlike other datasets, \textit{DeepFish} images capture a wide view of the underwater fish habitats.  The images also represent a diverse set of numerous habitats, and different underwater conditions.} Further, \textit{DeepFish} images are in-situ as they are extracted directly unaltered from the underwater camera. These images \green{can} also contain several fish that are potentially occluded and overlapping. In contrast, QUT images are post-processed. Most of the images in the QUT dataset are captured in "controlled" conditions, that is, the image collector spread the fish fins and captured the fish image against a constant background with controlled illumination then annotated all the images by drawing a tight red bounding box around the fish body. Fish4Knowledge and Rockfish images are taken in the fish natural habitat but they are also post-processed as they are cropped to ensure fish are at the center of the image (see Figure~\ref{qutf4k} for a comparison between the images from each dataset). Thus, \textit{DeepFish} is more suitable for training models for the purpose of analyzing fish in the wild, and it requires less effort for collecting additional images and annotations.}

\blue{The task that the other datasets address is limited to classification where  the goal is to distinguish between fish species. Fish4Knowledge and Rockfish also address the task of detection where the goal is to draw a bounding box around the fish. On the other hand, \textit{DeepFish} addresses 4 tasks, which are classification, counting, localization, and segmentation, which means algorithms that score well on this benchmark should be able to provide a comprehensive \green{analysis for the fish community}. Overall, the \textit{DeepFish} dataset exceeds previous fish datasets in terms of size, annotation richness, and scene complexity and variability.}

\begin{table}[t]
\centering
\caption{\blue{\textbf{Comparison between dataset characteristics}.  Clf, Cnt, Loc, Seg refer to the tasks of classification, counting, localization and segmentation. "in-situ" datasets consist of unaltered images that capture the fish underwater in their natural habitat, whereas  "controlled" consist of post-processed fish images where  background and illumination have been altered. "in-situ (cropped at center)" datasets have images cropped at the center where the fish is.
}}
\label{tab:datasets}
\resizebox{\textwidth}{!}{%
\begin{tabular}{l|c|c|c|c|c|c|c}
\textbf{Dataset} & \textbf{\# images} & \textbf{Tasks}                                                                                  & \textbf{\# fish/ image} & \textbf{Resolution} & \textbf{\# habitats} & \textbf{Environment type}                                              & \textbf{Has background images} \\\hline\hline
DeepFish (Ours)             & 39766              & \begin{tabular}[c]{@{}l@{}}Clf, Cnt, Loc, Seg \end{tabular} & $\sim$7                  & 1920 x 1080           & 20                   & in-situ                                                                & Yes                                        \\ \hline
QUT~\cite{anantharajah2014local}              & 3960               & Clf                                                                                  & 1                        & 480 x 360           & N/A                  & \begin{tabular}[c]{@{}l@{}}30\% in-situ, 70\% controlled\end{tabular} & No                                         \\ \hline
Fish4Knowledge~\cite{f4kFinalReport}   & 27370              & Detection                                                                                       & $\sim$1                        & 352 x 240           & N/A                  & in-situ (cropped at center)                                                                & No                                         \\ \hline
Rockfish~\cite{Rockfish2013}        & 4307               & \begin{tabular}[c]{@{}l@{}}Detection\end{tabular}                              & $\sim$1                  & 1280 x 720            & N/A                  & in-situ (cropped at center)                                                             & Yes                                        \\ \hline
\end{tabular}%
}
\end{table}

\section{Methods and Experiments}
\label{sec:method}

Based on the labels of \textit{DeepFish}, we consider these four computer vision tasks: classification, counting, localization, and segmentation. Deep learning have consistently achieved state-of-the-art results on these tasks as they can leverage the enormous size of the datasets they are trained on. These datasets include  ImageNet~\cite{imagenet_cvpr09}, Pascal~\cite{everingham2015pascal}, CityScapes~\cite{cordts2016cityscapes} and COCO~\cite{lin2014microsoft}. \textit{DeepFish} aims to be part of these large scale datasets with the unique goal of understanding complex fish habitats for the purpose of inspiring further research in this area.

We present standard deep learning methods for each of these tasks. Shown as the blue module in Figure~\ref{fig:models}, these methods have the ResNet-50~\cite{He_2016} backbone which is one of the most popular feature extractors for image understanding and visual recognition. They enable models to learn from large datasets and transfer the acquired knowledge to train efficiently on another dataset. This process is known as transfer learning and has been consistently used in most current deep learning methods~\cite{laradji2020looc}. Such pretrained models can even recognize object classes that they have never been trained on~\cite{rodriguez2020embedding}. This property illustrates how powerful the extracted features  are from a pretrained ResNet-50. 

Therefore, we initialize the weights of our ResNet-50 backbones by pre-training it on ImageNet following the procedure discussed in \citet{imagenet_cvpr09}. ImageNet consists of over 14 million images categorized over 1000 classes. As a result, the backbone learns to extract strong, general features for unseen images by training on such dataset. These features are then used by a designated module to perform their respective computer vision task such as classification and segmentation. We describe these modules in the sections below.

To put the results into perspective, we also include baseline results by training the same methods without ImageNet pretraining (Table~\ref{tab:results}). In this case, we randomly initialize the weights of the ResNet-50 backbone with  Xavier's method~\cite{Glorot2010UnderstandingTD}. These results also illustrate the efficacy of having pretrained models over randomly initialized models.

\begin{table}[t]
\caption{Comparison between randomly initialized and ImageNet pretrained models. Classification results were evaluated on the \textit{FishClf} dataset, counting and localization on the \textit{FishLoc} dataset, and segmentation on the \textit{FishSeg} dataset.}
\label{tab:results}
\centering
\begin{tabular}{l|c|c|ccc}
                   & \textbf{Classification} & \textbf{Counting} & \multicolumn{2}{c|}{\textbf{Localization}} & \multicolumn{1}{l}{\textbf{Segmentation}} \\
                   & Accuracy       & MAE      & MAE   & \multicolumn{1}{c|}{GAME} & mIoU                             \\ \hline
Random Weights     & 0.65           & 1.30     & 1.22  & \multicolumn{1}{c|}{1.30} & 0.49                             \\ \hline
Pretrained Weights & 0.99           & 0.38     & 0.21  & \multicolumn{1}{c|}{1.22} & 0.93                            
\end{tabular}
\end{table}

\subsection{Classification Results} 
The goal of the classification task is to  identify whether images are foreground (contains fish) or background (contains no fish). We use accuracy to evaluate the models on this task which is a standard metric for binary classification problems~\cite{french2015convolutional, LifeCLEF2014, BOOM2014, f4kFinalReport,mounsaveng2020learning}. The metric is computed as $$ACC=(TP+TN)/N,$$ where $TP$ and $TN$ are the true positives and true negatives, respectively, and $N$ is the total number of images. A true positive represents an image with at least one fish that is predicted as foreground, whereas a true negative represents an image with no fish  that is predicted as background. For this task we used the FishClf dataset for this task where the number of images labeled is 39,766. 

\begin{figure}[t]
\centering
\includegraphics[width=0.8\textwidth]{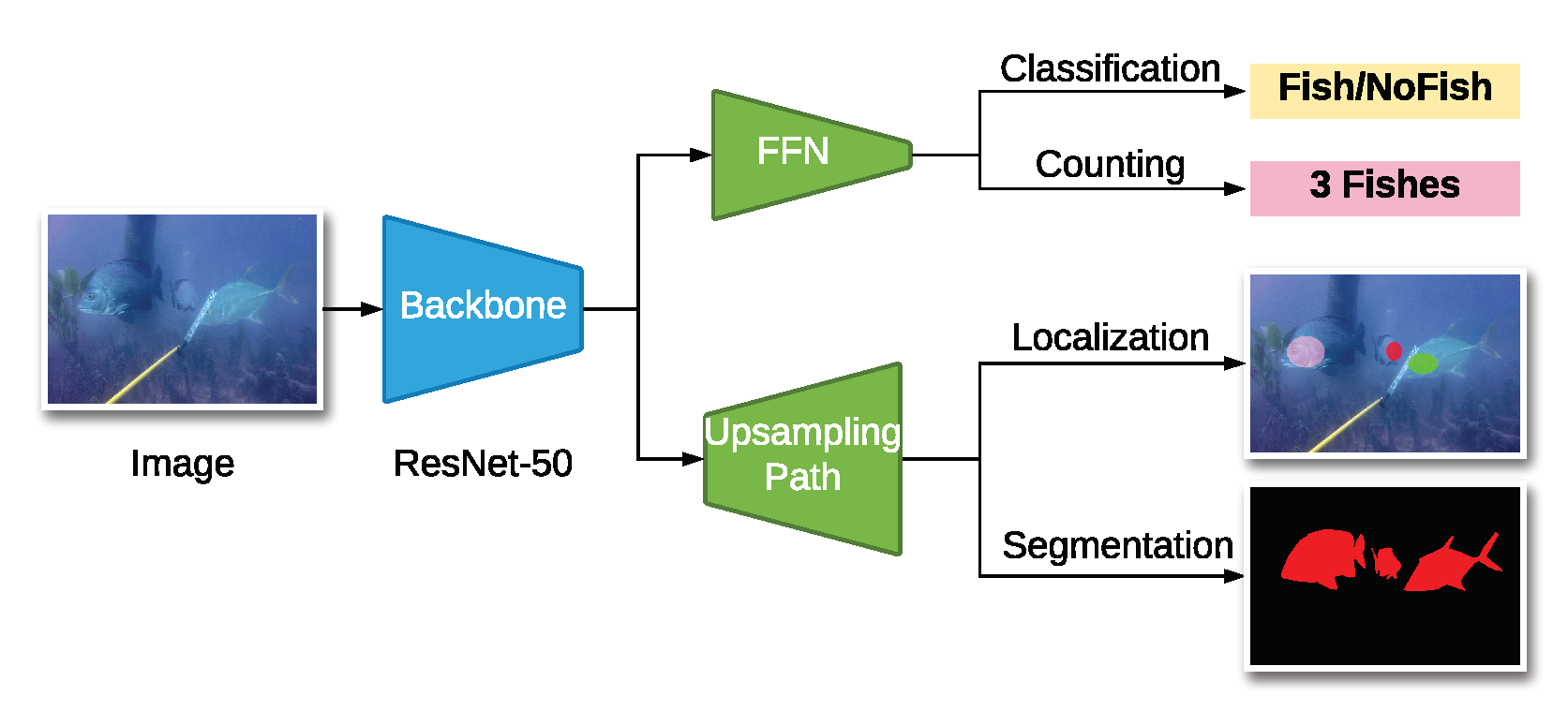}
\caption{\textbf{Deep learning methods.} The architecture used for the four computer vision tasks of classification, counting, localization, and segmentation consists of two components. The first component is the ResNet-50 backbone which is used to extract features from the input image. The second component is either a feed-forward network that outputs a scalar value for the input image or an upsampling path that outputs a value for each pixel in the image.}
\label{fig:models}
\end{figure}

The classification architecture consists of a ResNet-50 backbone and a feed-forward network (FFN) (classification branch of Figure~\ref{fig:models}). FFN takes as input features extracted by ResNet-50 and outputs a probability for the image corresponding to how likely it contains a fish. If the probability is higher than 0.5 the the predicted classification label is foreground. For the FFN, we use the network presented in ImageNet which consists of 3 layers. However, instead of the original 1000-class output layer, we use a 2-class output layer to represent the foreground or background class.

During training, the classifier learns to minimize the binary cross-entropy objective function~\cite{Murphy2012MachineL} using the Adam~\cite{Kingma2014AdamAM} optimizer. The learning rate was set as $10^{-3}$ and the batch size was set to be 16. Since FFN require a fixed resolution of the extracted features, the input images are resized to $224\times224$. At test time, the model outputs a score for each of the two classes for a given unseen image. The predicted class for that image is the class with the higher score.

In Table~\ref{tab:results} we compare between a classifier with the backbone pretrained on ImageNet and with the randomly initialized backbone. Note that both classifiers have their FFN network initialized at random. We see that the pretrained model achieved near-perfect classification results outperforming the baseline significantly. This result suggests that transfer learning is important and that deep learning has strong potential for analyzing fish habitats. 

\subsection{Counting Results}
The goal of the counting task is to predict the number of fish present in an image. We evaluate the models on the FishLoc dataset, which consists of 3200 images labeled with point-level annotations. We measure the model's efficacy in predicting the fish count by using the mean absolute error. It is defined as, $$ MAE=\frac{1}{N}\sum_{i=1}^N|\hat{C}_i-C_i|, $$
where $C_i$ is the true fish count for image $i$ and $\hat{C}_i$ is the  model's predicted fish count for image $i$. This metric is standard for object counting~\cite{Trancos2015,lempitsky2010learning} and it measures the number of miscounts the model is making on average across the test images.

The counting branch in Figure~\ref{fig:models} shows the architecture used for the counting task, which, similar to the classifier, consists of a ResNet-50 backbone and a feed-forward network (FFN). Given the extracted features from the backbone for an input image, the FFN outputs a number that correspond to the count of the fish in the image. Thus, instead of a 2-class output layer like with the classifier, the counting model has a single node output layer.

We train the models \blue{by minimizing} the squared error loss~\cite{Murphy2012MachineL}, which is a common objective function for the counting task. At test time, the predicted value for an image is the predicted object count. 

The counting model with the backbone pretrained on ImageNet achieved an MAE of 0.38 (Table~\ref{tab:results}. This result corresponds to making an average of 0.38 fish miscounts per image which is satisfactory as the average number of fish per image is 7. In comparison, the counting model initialized randomly achieved an MAE of 1.30. This result further confirms that transfer learning and deep learning can successfully address the counting task despite the fact that the dataset for counting (FishLoc) is much smaller than classification (FishClf).

\subsection{Localization Results}
Localization considers the task of identifying the locations of the fish in the image. It is a more difficult task than classification and counting as the fish can extensively overlap. Like with the counting task, we evaluate the models on the FishLoc dataset. However, MAE scores do not provide how well the model performs at localization as the model can count the wrong objects and still achieve perfect score. To address this limitation, we use a more accurate evaluation for localization by following~\cite{Trancos2015}, which considers both the object count and the location estimated for the objects. This metric is called Grid Average Mean absolute Error (GAME). It is computed as $$GAME = \sum_{i=1}^4 GAME(L),\;\;\;\;\;\;\;\;\;\;\;GAME(L) = \frac{1}{N}\sum_{i=1}^N(\sum_{l=1}^{4^L}|D^l_i - \hat{D}^l_i|),$$ where $D^l_i$ is the number of point-level annotations in region $l$, and $\hat{D}^l_i$ is the model's predicted count for region $l$. $GAME(L)$ first divides the image into a grid of $4^L$ non-overlapping regions, and then computes the sum of the MAE scores across these regions. The higher $L$, the more restrictive the GAME metric will be. Note that $GAME(0)$ is equivalent to MAE.

The localization branch in Figure~\ref{fig:models} shows the architecture used for the localization task, which consists of a ResNet-50 backbone and an upsampling path. The upsampling path is based on the network described in FCN8~\cite{long2015fully} which is a standard fully convolutional neural network meant for localization and segmentation, which consists of 3 upsampling layers.

FCN8 processes images as follows. The features extracted with the backbone are of a smaller resolution than the input image. These features are then upsampled with the upsampling path to match the resolution of the input image. The final output is a per-pixel  probability map where each pixel represents the likelihood that it belongs to the \textit{fish} class. 

The models is trained using a state-of-the-art localization-based loss function called LCFCN~\cite{Laradji2018ECCV}.  LCFCN is trained using 4 objective functions: image-level loss, point-level loss, split-level loss, and false positive loss. The image-level loss encourages the model to \blue{predict all pixels as background for background images}. The point-level loss encourages the model to predict the centroids of the fish. Unfortunately, these two loss terms alone do not prevent the model from predicting every pixel as fish for foreground images. Thus, LCFCN also minimizes the split loss and false-positive loss. The split loss splits the predicted regions so that no region has more than one point annotation. This results in one blob per point annotation. The false-positive loss prevents the model from predicting blobs for regions where there are no point annotations. Note that training LCFCN only requires point-level annotations which are spatial locations of where the objects are in the image.

\begin{figure}[t]
  \centering
\includegraphics[width=\textwidth]{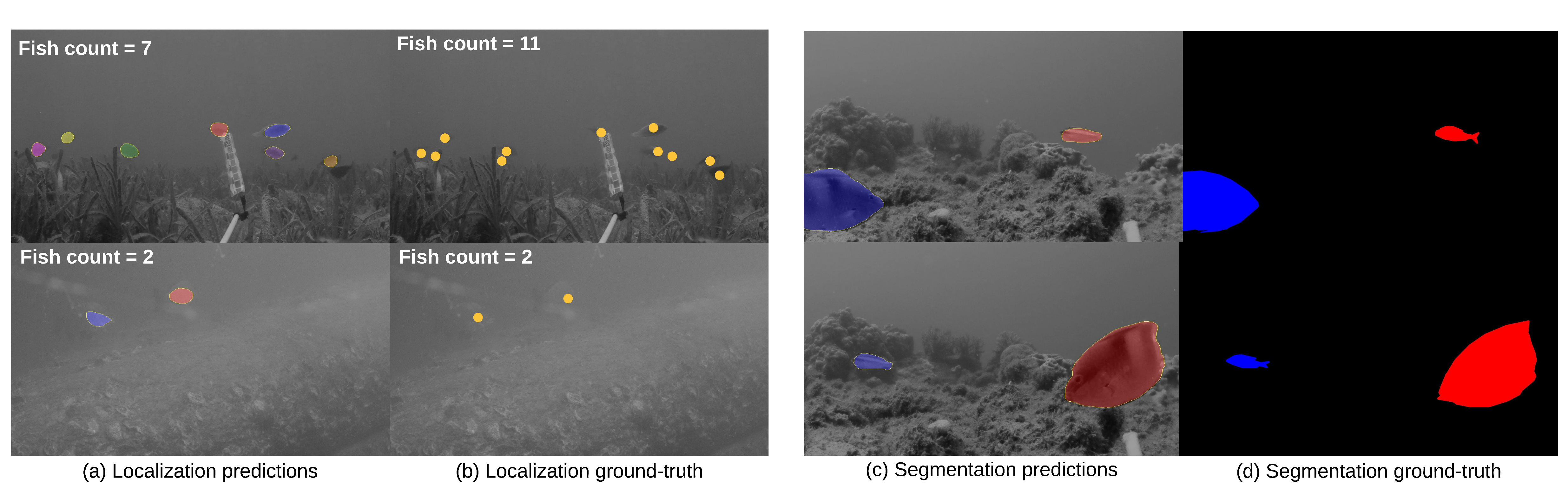}
\caption{Qualitative results on counting, localization, and segmentation. 
 (a) Prediction results of the model trained with the LCFCN loss~\cite{Laradji2018ECCV}. (b) Annotations that represent
the $(x,y)$ coordinates of each fish within the images. (c) Prediction results of the model trained with the focal loss~\cite{lin2017focal}. (d) Annotations that represent the full segmentation masks of the corresponding fish.}
\label{fig:qualitative}
\end{figure}

At test time, the predicted probability map are thresholded to become 1 if they are larger than 0.5 and 0 otherwise. This results in a binary mask, where each blob is a single connected component and they can be collectively obtained using the standard connected components algorithm. The number of connected components is the object count and each blob represents the location of an object instance (see Figure~\ref{fig:qualitative} for example predictions with FCN8 trained with LCFCN).

Models trained on this dataset are optimized using Adam~\cite{Kingma2014AdamAM} with a learning rate of $10^{-3}$ and weight decay of $0.0005$, and \blue{have been ran} for 1000 epochs on the training set. In all cases the batch size is 1, which makes it applicable for machines with limited memory. 

Table~\ref{tab:results} shows the MAE and GAME results of training an FCN8 with and without a pretrained ResNet-50 backbone using the LCFCN loss function. We see that pretraining leads to significant improvement on MAE and a slight improvement for GAME. The efficacy of the pretrained model is further confirmed by the qualitative results shown in Figure~\ref{fig:qualitative}(a) where the predicted blobs are well-placed on top of the fish in the images.

\subsection{Segmentation Results}
The task of segmentation is to label every pixel in the image as either fish or not fish (Figure~\ref{fig:qualitative}(c-d)). \green{When combined } with depth information, a segmented image allows us to measure the size and the weight of the fish in a location, which can vastly improve our understanding of fish communities. We evaluate the model on the FishSeg dataset for which we acquired per-pixel labels for 620 images. We evaluate the models on this dataset using the standard Jaccard index~\cite{everingham2015pascal, cordts2016cityscapes} which is defined as the number of correctly labelled pixels of a class, divided by the number of pixels labelled with that class in either the ground truth mask or the predicted mask. It is commonly known as the intersection-over-union metric IoU, computed as $\frac{TP}{TP + FP + FN}$, where TP, FP, and FN are the numbers of true positive, false positive, and false negative pixels, respectively, which is determined over the whole test set. In segmentation tasks, the IoU is preferred over accuracy as it is not as affected by the class imbalances that are inherent in foreground and background segmentation masks like in \textit{DeepFish}.

During training, instead of minimizing the standard per-pixel cross-entropy loss~\cite{long2015fully}, we use the focal loss function~\cite{lin2017focal} which is more suitable when the number of background pixels is much higher than the foreground pixels like in our dataset. The rest of the training procedure is the same as with the methods trained for localization.

At test time, the model outputs a probability for each pixel in the image. If the probability is higher than 0.5 for the foreground class, then the pixel is labeled as fish, resulting in a segmentation mask for the input image.

The results in Table~\ref{tab:results} show a comparison between the pretrained and randomly initialized segmentation model. Like with the other tasks, the pretrained model achieves superior results both quantitatively and qualitatively (Figure~\ref{fig:qualitative}).

\section{Conclusions and Perspectives}
We have introduced \textit{DeepFish} as a benchmark suite consisting of a large-scale dataset for the purpose of developing new models that can efficiently analyze remote underwater fish habitats. Compared to current fish datasets, \textit{DeepFish} consists of a diverse set of images that capture complex scenes from a large set of fish habitats that span coastal marine-environments of tropical Australia. We acquired point-level and per-pixel annotations and designed experimental setups that enable models to be evaluated for the tasks of classification, counting, localization and segmentation. We also present results demonstrating the efficacy of standard deep learning methods that were pretrained on ImageNet. These results can be used as baseline to help evaluate new models for this problem setup.

For future work, we plan to adapt \textit{DeepFish} by adding new benchmarks and annotations in order to inspire fish analysis models for other useful use cases. Thus, we will consider challenges that fall under weak supervision, active learning, or few-shot learning where the goal is to train on datasets whose labels were collected with minimal human effort.

\section*{Data Availability}
\label{sec:Data_Availability}

The {\it DeepFish} dataset and the code is publicly available at \url{https://alzayats.github.io/DeepFish/} and \url{https://github.com/alzayats/DeepFish}, respectively.

\bibliography{references}

\section*{Acknowledgements}
This research is supported by an Australian Research Training Program (RTP) Scholarship. We gratefully acknowledge Strategic Research Initiative Funding (SRIF-2018) of James Cook University. Issam Laradji is funded by the UBC Four-Year Doctoral Fellowships (4YF).

\section*{Author Contributions}
A.S. and I.H.L. contributed equally to writing the manuscript, coding and annotating the dataset. D.V. reviewed, revised the manuscript and conceived the project. D.A.K. revised the manuscript. M.B. and M.S collected the dataset and reviewed the manuscript.

\paragraph{Corresponding author}:\\Correspondence to alzayat.saleh@my.jcu.edu.au

\section*{Additional Information}
\textbf{Competing interests} The authors declare no competing interests.\\
\textbf{Ethical approval} This work was conducted with the approval of the JCU Animal Ethics Committee (protocol A2258), and conducted in accordance with DAFF general fisheries permit \#168652 and GBRMP permit \#CMES63.

\end{document}